\documentclass[preprint,12pt,authoryear]{elsarticle}

\usepackage{amsmath,amssymb,amsfonts}
\usepackage{tabularx} 
\usepackage{algorithm}
\usepackage{algorithmic}
\usepackage{graphicx}
\usepackage{textcomp}
\usepackage{xcolor}
\usepackage{url}

\journal{Energy and Buildings}

\begin{document}

\begin{frontmatter}



\title{A Transfer Learning Approach to Minimize Reinforcement Learning Risks in Energy Optimization for Smart Buildings}


\author[1]{Mikhail Genkin}
\author[1]{J.J. McArthur}

\affiliation[1]{organization={Department of Architectural Science},
            addressline={Toronto Metropolitan University}, 
            city={Toronto}, 
            state={Ontario},
            country={Canada}}

\begin{abstract}
Energy optimization leveraging artificially intelligent algorithms has been proven effective. However, when buildings are commissioned, there is no historical data that could be used to train these algorithms. On-line Reinforcement Learning (RL) algorithms have shown significant promise, but their deployment carries a significant risk, because as the RL agent initially explores its action space it could cause significant discomfort to the building residents. In this paper we present ReLBOT – a new technique that uses transfer learning in conjunction with deep RL to transfer knowledge from an existing, optimized and instrumented building, to the newly commissioning smart building, to reduce the adverse impact of the reinforcement learning agent's warm-up period. We demonstrate improvements of up to 6.2 times in the duration, and up to 132 times in prediction variance, for the reinforcement learning agent's warm-up period.
\end{abstract}

\begin{graphicalabstract}
\includegraphics[width=3.5in]{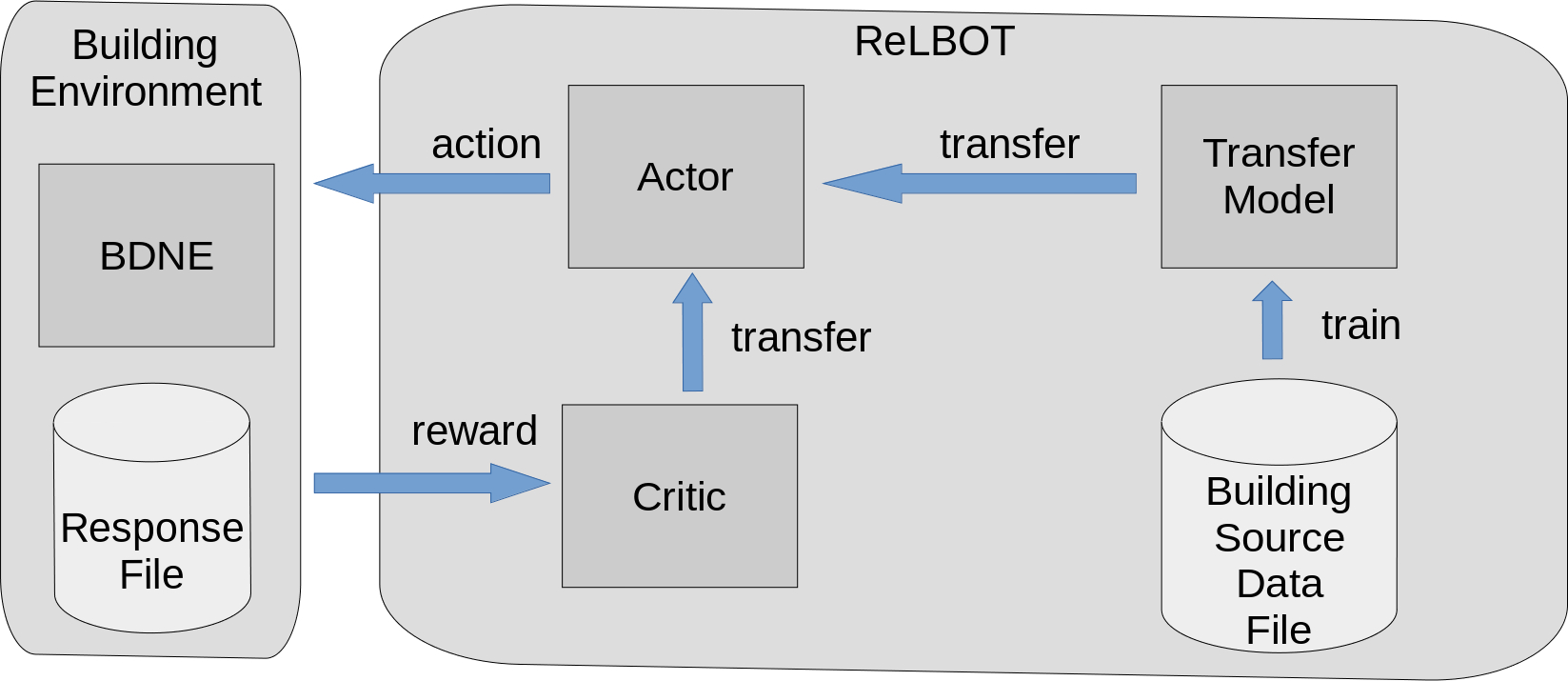}
\par
\par
\includegraphics[width=3.5in]{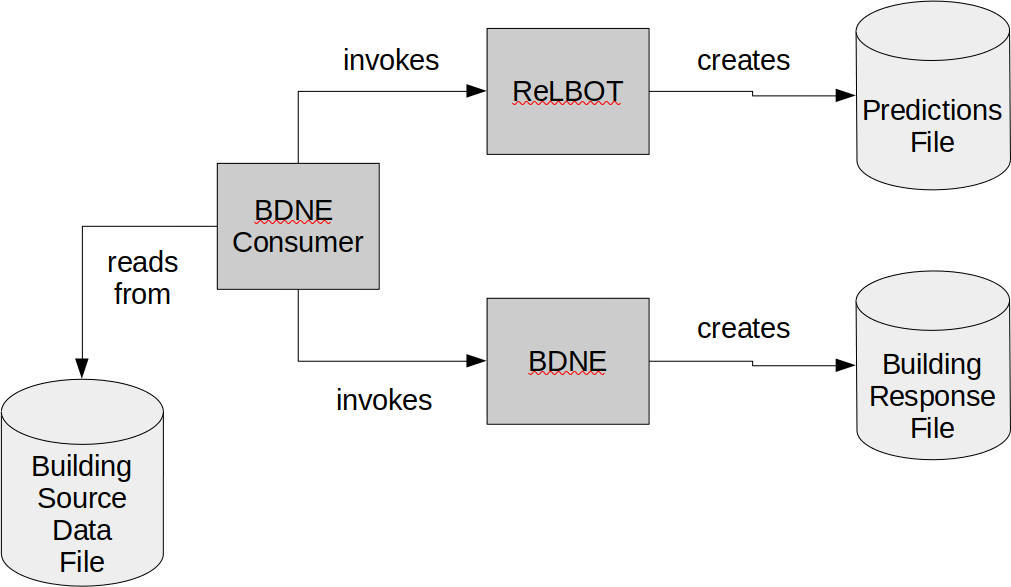}
\par
\par
\includegraphics[width=3.5in]{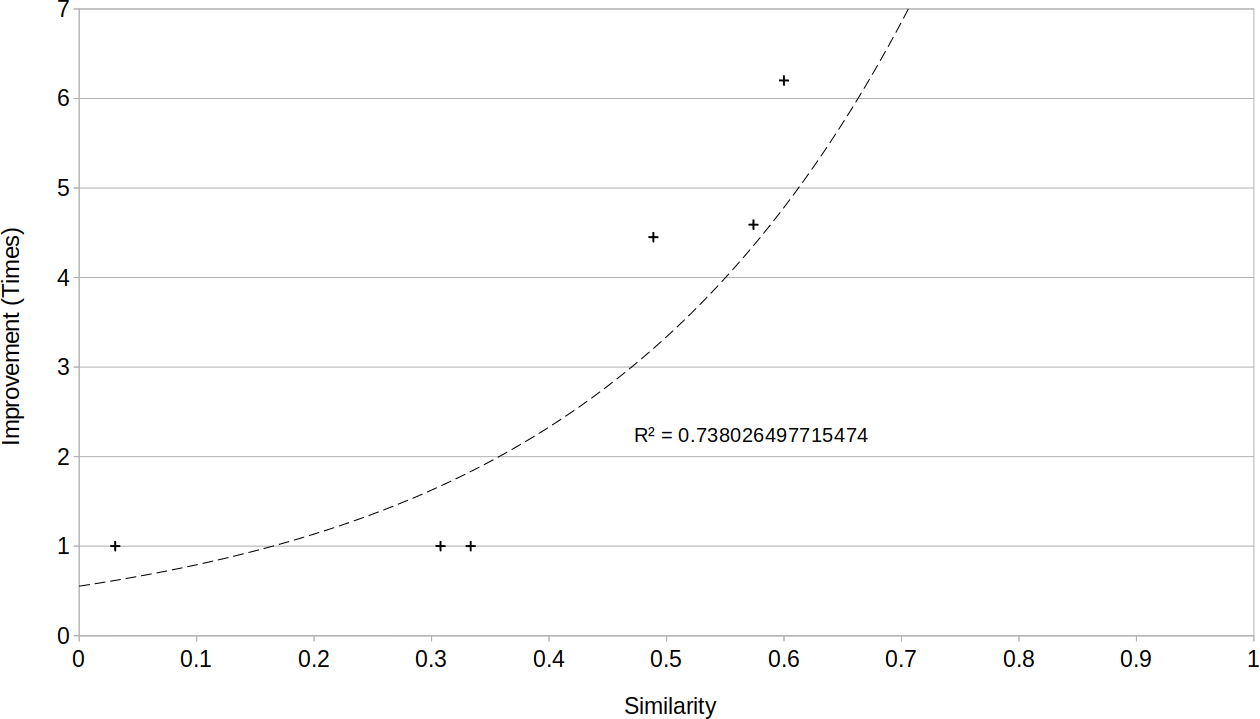}
\end{graphicalabstract}

\begin{highlights}
\item The first similarity-informed transfer learning method to implement reinforcement learning for building energy optimization.
\item Warm-up period duration reduced by up to 6.2 times.
\item Prediction variance reduced by up to 132 times.
\end{highlights}


\end{frontmatter}



\section{Introduction} \label{introduction}

\subsection{Smart Buildings and Energy Optimization} \label{intro-energy-opt}

Buildings are responsible for approximately 32\% of global energy use and 19\% of $CO_2$ emissions; further, their longevity as well as the ability to reduce these values have made it a significant priority for emissions reduction \citep{IPCC2018}. In this context, the ability to manage buildings most efficiently is critical. In temperate climate, the majority of building energy consumption is due to Heating, Ventilation, and Air Conditioning (HVAC) loads. HVAC systems are critical to ensure a healthy and comfortable indoor environment for the building occupants. Chillers (cooling) and boilers (heating) are the most significant HVAC equipment and the ability to optimize their controls offers significant potential for $CO_2$ emissions reduction.

Recent developments in Machine Learning (ML) and cloud computing provide new opportunities for controls optimization. A significant number of studies have explored online controls optimization, demonstrating savings of 30-70\% \citep{kannan2020ac, teo2021energy, stocketal2021}. However, there is limited uptake due to research gaps in the development of  data retrieval, analysis, and management processes; services for deployment, maintenance, and calibration of sensors; the high cost of updating physics-based models for performance and energy estimation; inability to scale machine learning algorithms for building energy management; and the lack of case studies \citep{minolietal2017}. Further underpinning the process challenges is a lack of a supporting computational architecture to integrate ML and Artificial Intelligence (AI) with building systems \citep{genkin2023b}. Reinforcement Learning (RL) offers significant opportunity to streamline the process, reducing the need for physics-based models \citep{vazquez2019reinforcement}. However, they are challenging to implement due to a) the time-consuming and data-demanding training process; b) the need to ensure the RL agent will not create problems during its training stage; and c) the lack of knowledge how to implement Transfer Learning (TL) to minimize the risks of b) \citep{wang2020reinforcement}. This paper directly addresses this last challenge, presenting a novel means to integrate RL and TL for building energy optimization.

\subsection{Problem} \label{problem}

The problem with using off-line reinforcement learning, or any machine learning technique that relies on data collected in the past to optimize performance in the present is that it can take a very long time to collect a sufficient volume of data to enable acceptable algorithm performance. Building life cycles are measured in decades. Seasonal variations in external temperature, humidity, and other environmental factors affect building energy performance. When a new smart building is commissioned, or a traditional building is converted to be a smart building, years may pass before a sufficient volume of data, capturing seasonal building performance under a variety of external conditions, has been collected to enable effective application of these machine learning algorithms. This problem is referred to as the smart building 'cold start'. On-line reinforcement learning methods have been applied to deal with this issue. However, an on-line reinforcement learning agent learns by trial-and-error. During it's initial period of operation, known as the warm-up period, the agent's actions are largely random, and could cause significant discomfort to the building's residents because these actions involve adjusting chiller and boiler set-points, which in turn, results in changes to the building's internal air temperature. Deployment of reinforcement learning algorithms thus carries significant risk for building management companies, and this consideration has inhibited the adoption of on-line reinforcement learning algorithms. 

\subsection{Contribution} \label{contribution}

In this paper we present Reinforcement Learning Building Optimizer with Transfer learning (ReLBOT) - the first technique specifically designed to overcome the smart building cold start. Our technique uses deep reinforcement learning in combination with transfer learning to enable algorithm training on an established building with a sufficient level of instrumentation and a historical data set, and subsequent application of this algorithm on a different, brand new, smart building that lacks historical data. We also present a method for measuring similarity among buildings and selecting the most appropriate donor building to optimize transfer learning. Our approach allows the close-to-optimal energy performance for the building to be established at the very beginning of the building's life-cycle, without causing discomfort to the building residents.

\section{Previous Work} \label{prev-work}

To contextualize this research, we provide a review of two bodies of literature: those related specifically to overcoming the ‘cold start’ problem, both within the building energy domain and the broader literature, and those applying ML algorithms to optimize building energy performance. We begin by recapping research that describes how the smart building cold start problem was dealt with up until this point and expand this with perspectives from the broader literature to provide a more holistic context to the issue. We then proceed to discuss the most relevant prior art focusing on the use of various ML algorithms to optimize building energy performance, and then we conclude our review by highlighting the potential to TL to overcome the ‘cold start’ problem with RL algorithms and examine the few domain-specific works that address this combination of approaches.

\subsection{Overcoming Cold Start} \label{prev-cold-start}

The ‘cold start’ problem for building energy simulation is a long-recognized problem for data science, particularly recommender systems \citep{scheinetal2002}.  In this context, ‘cold start’ refers to the challenge of predicting energy consumption for a new facility for which no historical data is available \citep{chadoulos2021}; this has been noted as a significant challenge to be overcome to develop building energy-efficiency recommender systems \citep{himeur2021survey}. To understand the state of the art in terms of how to overcome this challenge in energy optimization, we begin with a review of that literature and then expand beyond the building energy domain to obtain a more holistic view.

The cold start problem with respect to HVAC optimization has been addressed through data augmentation by a number of scholars, notably Kannan et al. \citep{kannan2020ac, kannan2020artificial}. In one study,  \cite{kannan2020artificial} used preference maps to expand the dataset for air-conditioners (ACs), adding similar data from other ACs to help overcome the cold-start problem and leading to a common deep neural network for all units. Considering a large number of ACs (37,748), a large dataset was created despite the relatively small number of points for each, and the resultant model showed excellent results with a median 57.38\% achieved energy savings. In a related study, \cite{kannan2020ac} considered a large dataset of 53,528 ACs, finding that individual model predictions were insufficient in 76\% of cases, while a combined dataset and deep neural network permitted strong (R2= 0.8) predictions to be made for all ACs.  Similarly, \cite{chadoulos2021} used an aggregate dataset to train a common deep learning model (recurrent neural network encoder with multi-layer perceptron architecture) for household energy demand forecasting, significantly increasing predictive accuracy over single-house models. This data augmentation approach is was also implemented by \cite{weietal2020}, who applied Markov decision process with deep reinforcement learning to develop a recommender system to guide occupant energy-conservation behavior, achieving savings of 19-26\% across their field studies. Data augmentation has also been used for electric load prediction using tree-based (multivariate random forest) methods \citep{moon2020solving}.

Bridging between building energy and other disciplines, \cite{salinasetal2020} explored the application of deep learning to time series forecasting using auto regressive recurrent networks, demonstrating the value of this approach to both traffic and electricity consumption forecasting. Of particular interest to energy optimization, this approach was able to incorporate seasonality to generate high-accuracy forecasts.

There have been recent attempts to mitigate the cold start problem using rules-based control algorithms \citep{LU2023112854, NWEYE2023121323}. These algorithms can produce reasonable energy performance during the cold start period. However, the energy performance of the building will not be optimal. To be certain that the building energy performance is optimal, a search of the building parameter space needs to be performed, and this can cause discomfort to the building residents.

Beyond the building energy domain, a significant volume of research has also considered the ‘cold start’ issue for recommender systems. Collaborative filtering is the most common approach for recommender systems \citep{wei2017collaborative}, however, it suffers extensively from the ‘cold start’ issue \citep{fu2019deeply, wei2017collaborative, himeur2021survey}. To overcome this, deep learning has been increasingly used to use cross-domain information to help overcome the cold start approach. Studies of this kind have explored deep neural network integration with collaborative filtering \citep{kiran2020dnnrec} and stacked  autoencoders \citep{fu2019deeply}. Other studies seeking to overcome the cold start problem have explored content-based and context-aware approaches such as hybrid collaborative filtering and content-based approaches \citep{ojagh2020location}, influential context embedding and neural context-aware units \citep{hu2019hers}, and contextual bandit reinforcement learning \citep{wanigasekara2016bandit}.

\subsection{ML and Smart Buildings} \label{prev-ml-sb}

There is an extensive body of literature on the application of ML to Smart Building systems; for comprehensive reviews on these applications and their supporting software and architectures,  refer to \citep{quolomanyetal2019, minolietal2017} and for insight on AI-initiated learning  in these applications, refer to \citep{alannesierla2022}.

ML is used for a range of tasks to support buildings, including: data acquisition, data pre-processing, feature extraction, selection, and prediction, and dimension reduction \citep{quolomanyetal2019}. Within the HVAC domain, building automation systems (BAS) have proven to be a valuable source of data alongside additional energy metering and, in some cases, supplemental equipment controls points \citep{minoli2017iot, stocketal2021} and PoE devices \citep{minoli2017iot}. Other building management applications of ML include lighting \citep{huang2018}, water management \citep{vrsalovic2021}, energy management \citep{minolietal2017}, indoor environmental control \citep{zhangetal2020, ascioneetal2016}, automated fault detection \citep{mirnaghi2020fault}, and occupant detection \citep{zhaoetal2018}.

Of particular relevance to this paper is the application of ML to HVAC controls and energy management. Numerous studies have shown significant potential for energy management \citep{minolietal2017}, with demonstrated savings exceeding 50\%  \citep{ascioneetal2016, stocketal2021}.  Such applications have leveraged the full spectrum of ML techniques \cite{alannesierla2022, minolietal2017, quolomanyetal2019}. Supervised learning (classification, regression, ensemble methods, and time-series analysis) has been widely used, but requires a significant volume of labelled data \citep{minolietal2017, quolomanyetal2019}, which makes them highly susceptible to the ‘cold start’ problem described above for both energy management and fault detection applications. Unsupervised learning, primarily using clustering, overcomes this challenge but is computationally expensive and has limited applications \citep{quolomanyetal2019}. Semi-supervised learning overcomes some of these challenges but has had limited adoption for energy applications. Finally, RL has had increasing adoption for energy management and HVAC control \citep{quolomanyetal2019, wang2020reinforcement, alannesierla2022}, but suffers from a lack of real-world application due to the risks present during agent training \citep{wang2020reinforcement}, the high computational cost control \citep{quolomanyetal2019, wang2020reinforcement}, and the need for significant training data for robust operation, again making it susceptible to the ‘cold start’ problem \citep{wang2020reinforcement}.

TL has been identified as offering significant promise to resolve the ‘cold start’ issue, but the application of TL to RL is recognized as a significant gap requiring additional research \citep{wang2020reinforcement, alannesierla2022} \cite{zhu2020transfer} provide a robust discussion of the theoretical integration of transfer and reinforcement learning, however there is a paucity of studies exploring this in the building energy domain.  The rise in application of TL for building energy prediction has been noted – see, for example, the review by \cite{himeur2022next} or papers by \cite{ribeiro2018transfer}, \cite{grubinger2017generalized}, and \cite{mocanu2016unsupervised} - however the majority of noted studies do not consider it in combination with RL \citep{deng2021reinforcement}. Four exceptions were noted. The first paper \citep{mocanu2016unsupervised} was limited to energy prediction only, developing an initial model based on a Deep Belief Network for transfer to other buildings of both the same and different occupancy types to predict energy consumption with RL. More recently, \cite{xuetal2020} sought to optimize HVAC controllers using deep reinforcement learning by decomposing their HVAC control system into a transferable front-end network with a building-agnostic back-end network, finding that this approach was effective even when the source and target buildings had differing numbers of thermal zones, materials and layouts, HVAC equipment, and – in some cases – weather conditions. Two other recent studies explored deep reinforcement learning with transfer learning. One \citep{fang2023cross} applied deep Q-learning to create a control strategy for the the source building, transferring the first few layers to the target building and refining the remaining layers using target building data. The second \citep{coraci2023online} used an online transfer learning (OTL) strategy to transfer a deep reinforcement learning (DRL) control policy based on a soft actor-critical approach. Both studies reported positive results, highlighting the significant potential for TL to transfer RL control strategies between buildings.  This paper contributes to this emerging discourse by presenting an architecture designed specifically to use TL to reduce the ‘cold start’ challenge for RL building optimization algorithms.

\section{Architecture} \label{architecture}

Results presented in this paper were generated using two separate systems:

\begin{enumerate}
\item The Building Data Neural Emulator (BDNE) was used to emulate the data behavior of an actual building.
\item The RelBOT itself.
\end{enumerate}

Below we begin by describing the theoretical basis for our technique and then proceed to discuss the architecture and operation of BDNE, and subsequently the architecture and operation of ReLBOT.

\subsection{ReLBOT Theoretical Basis} \label{relbot-theory}

Before delving into the ReLBOT architecture we clarify the terminology that will be used in the remainder of this work.  The term \textit{target building} will be used to indicate the building  that is being optimized by ReLBOT for energy performance. The term \textit{transfer building} will be used to indicate the building that is being used as the source of data for transfer learning.

The method described in this work involves transferring knowledge learned in the transfer building domain $\mathfrak{D}_s$ to the target building domain $\mathfrak{D}_t$. The state of transfer building is described by a feature vector $X_s = \{x_1,...,x_n\} \in \mathcal{X}_s$, the transfer building feature space. In all cases $x_i \in \mathbb{R}$.

The label space of the transfer building is denoted as $Y_s$. The predictive task, in the transfer building domain, involves learning an objective function $f_s: \mathcal{X}_s \rightarrow \mathcal{Y}_s$. The task $\mathcal{T}_s = \{\mathcal{Y}_s, f(X_s)\}$, is learned from training data $\{x_i,y_i\}$ where $x_i \in \mathcal{X}_s$ and $y_i \in \mathcal{Y}_s$. The regression task $\mathcal{T}_s$ aims to predict the building Coefficient Of Performance (COP) given the input feature vector $X_s$.

The COP of the transfer building (any building) can be calculated using the following equation:

$$ COP_i = y_i = {c_w{\rho}_w\mathcal{F}_{cps}f(T_{in} - T_{out})} \div {\mathcal{E}_{ch}}$$

Where $c_w$ is the specific heat capacity of water, ${\rho}_w$ is the density of water, $\mathcal{F}_{cps}$ is the chiller pump speed, $f$ is the flow rate factor, $f$ is the flow rate factor, $T_{in}$ is the entering chiller temperature, $T_{out}$ is the exiting chiller water temperature, and $\mathcal{E}_{ch}$ is the energy consumption rate of the chiller.

Thus, we can see that since historical data for the transfer building exist it is possible to construct the label space $\mathcal{Y}_s$ and train the predictive function $f(X_s)$ so that it is able to perform the regression task $\mathcal{T}_s$ with sufficient accuracy by using standard machine learning approaches. This does require that $X_s$ contain features with values for $\mathcal{F}_{cps}$, $T_{in}$, $T_{out}$, and $\mathcal{E}_{ch}$. The other terms in the equation are known constants.  

There are no historical training data that can be used for the target building. This is the fundamental constraint of the cold start scenario that this work aims to address. The ReLBOT aims to solve this problem using an actor-critic reinforcement learning agent (see Figure ~\ref{relbot-architecture}).

The state of target building is described by a feature vector $X_t = \{x_1,...,x_m\} \in \mathcal{X}_t$, the transfer building feature space. As with the transfer building, in all cases $x_i \in \mathbb{R}$. The label space of the target building in the domain $\mathfrak{D}_t$ is denoted as $\mathcal{Y}_t$.  

The ReLBOT actor performs a classification task ${\mathcal{T}_t}^a$. This task involves learning a function $a(X_t)$ that returns the most appropriate action based on the building state represented by $X_t$ and using this function to select the most appropriate action at each step. There are no label data ${Y_t}^a$ for this task, and so the actor shares knowledge with the critic to ensure prediction accuracy.

The predictive task, in the target building domain, involves learning an objective function for the critic ${f_t}^c: \mathcal{X}_t \rightarrow {\mathcal{Y}_t}^c$. The critic task ${\mathcal{T}_t}^c = \{{\mathcal{Y}_t}^c, {f_t}^c(X_t)\}$, is learned from training data $\{x_i,y_i\}$ where $x_i \in \mathcal{X}_t$ and $y_i \in {\mathcal{Y}_t}^c$. The labels ${\mathcal{Y}_t}^c$ for the critic are, just like for the transfer building transfer building label space $\mathcal{Y}_t$, are COP value calculated for each input vector $X_t$ using the same formula given above.

Thus, the critic task for the target building ${\mathcal{T}_s}^c$ is the same regression task as the regression task for the transfer building $\mathcal{T}_s$, and it should be possible to share knowledge:  $\mathcal{T}_s \Rightarrow {\mathcal{T}_t}^c$. The critic shares knowledge ${\omega_t}^c$ with the actor function $a(X_t)$.

It must be noted, however, that the dimension of the transfer building input vector $n$ is not the same as the dimension of the target building input vector $m$, and, therefore, an adaptation function $\alpha_t(X_s,X_t)$ must be introduced to reconcile this difference. 

The critic is initialized using knowledge ${\omega_s}$ produced by the target building regression task ${\mathcal{T}_s}$. The critic is then incrementally retrained after each time that the input vector $X_t$ is presented. Algorithm ~\ref{relbot-algorithm} describes the operation of the ReLBOT actor-critic reinforcement learning agent.

\begin{algorithm}
\caption{The ReLBOT main algorithm.}
\begin{algorithmic}
\REQUIRE ${\{X_t}\}_{t=1}^k \neq \emptyset$ \COMMENT{Time series containing target building feature vectors.}
\REQUIRE $\omega_s$ \COMMENT{Knowledge from the transfer building.}
\REQUIRE ${\{A\}^j_{a=1}}$ \COMMENT{The set of allowed actions.}
\ENSURE $argmin(R)$ \COMMENT{Minimize instantaneous reward.}
\STATE \COMMENT{Initialize critic knowledge.}
\FORALL{t}
	\STATE initialize $r_t = 0.0$
    \FORALL{a}
		\STATE predict the instantaneous reward $r^a_t$
		\IF{$r^a_t$ $>$ ${r^{a-1}}_t$}
			\STATE $r_t = r^a_t$
		\ENDIF
	\ENDFOR
	\STATE calculate the actual reward $R_t$
	\STATE $y_t = R_t$
	\STATE incrementally train the critic using ${\{Y_t}\}_{t=1}^k$
	\STATE ${\mathcal{T}_t}^c \Rightarrow {\mathcal{T}_t}^a$ \COMMENT {transfer knowledge from critic to actor.}
\ENDFOR
\end{algorithmic}
\label{relbot-algorithm}
\end{algorithm}

Effectiveness of the transfer learning approach among buildings may vary. It is logical to assume that relatively similar buildings will benefit from transfer learning more than relatively dissimilar ones. It is therefore important to define \textit{similarity} among buildings mathematically.

In this work similarity between the transfer building and the target building is defined as $S_s^t = s({\{X_s}\}_{t=i}^k, {\{X_t}\}_{t=j}^l)$.

$$S_s^t \in \mathbb{R} \mid 0 \leqq s({\{X_s}\}_{t=i}^k, {\{X_t}\}_{t=j}^l) \leqq 1$$

Using this definition stating that two buildings are completely dissimilar would imply that the buildings have completely dissimilar feature vectors and $S_s^t = 0$. Conversely, stating that two buildings are perfectly similar would imply that their feature vectors are perfectly similar, and that $S_s^t = 1$. 

In order to calculate similarity some historical data must be available for both the transfer (${\{X_s}\}_{t=i}^k$) and the target building (${\{X_t}\}_{t=j}^l$). These time-series do not need to be the same length, and it is understood that the target building will have less historical data available, but they should contain segments capturing similar seasonality (for example the summer months).

\begin{algorithm}
\caption{Algorithm for computing similarity between the transfer building and the target building.}
\begin{algorithmic}
\REQUIRE ${\{X_s}\}_{t=1}^k \neq \emptyset$ \COMMENT{Time series containing transfer building feature vectors.}
\REQUIRE ${\{X_t}\}_{t=1}^k \neq \emptyset$ \COMMENT{Time series containing target building feature vectors.}
\ENSURE $0 \leqq S_s^t \leqq 1$ \COMMENT{Return similarity value between 0 and 1.}
\STATE $\zeta = 0$
\STATE $\varepsilon = 0$
\FORALL{n}
	\STATE $k_n = Kurt[x_n]$
	\STATE $sk_n = Skew[x_n]$
    \FORALL{m}
    	\STATE $k_m = Kurt[x_n]$
		\STATE $sk_m = Skew[x_n]$
		\IF{$(k_n  \thicksim k_m) \wedge (sk_n \thicksim sk_m) \wedge (\mu_n \thicksim \mu_m)$}
			\STATE $\zeta = \zeta + 1$
		\ENDIF
	\ENDFOR
\ENDFOR
\IF{$n > m$}
	\STATE $\varepsilon = 1-m/n$	
\ENDIF
\STATE $S_s^t = \zeta/n - \varepsilon$
\end{algorithmic}
\label{similarity-algorithm}
\end{algorithm}

The algorithm for calculating $S_s^t$ is listed in Algorithm ~\ref{similarity-algorithm}. The algorithm takes example time-series collected for the transfer building and target building. The algorithm then iterates through all of the features in the transfer building feature vector and tries to find a similar feature in the target building feature vector. Features are considered similar if they have similar normalized kurtosis and skew (both positive or both negative), and means. The means are considered similar if the distance between them is less than the sum of the standard deviation for the transfer building feature and the standard deviation of the target building feature that is being considered.

If the length of the transfer building feature vector $n$ is greater than the length of the target building feature vector $m$, a penalty ($\varepsilon$ in Algorithm ~\ref{similarity-algorithm}) is applied. This is due to the consideration that it will not be possible to transfer all of the knowledge extracted from a richer feature vector. 

\subsection{BDNE Architecture} \label{bdss-architecture}

\begin{figure}[htbp]
\centerline{\includegraphics[width=3.5in]{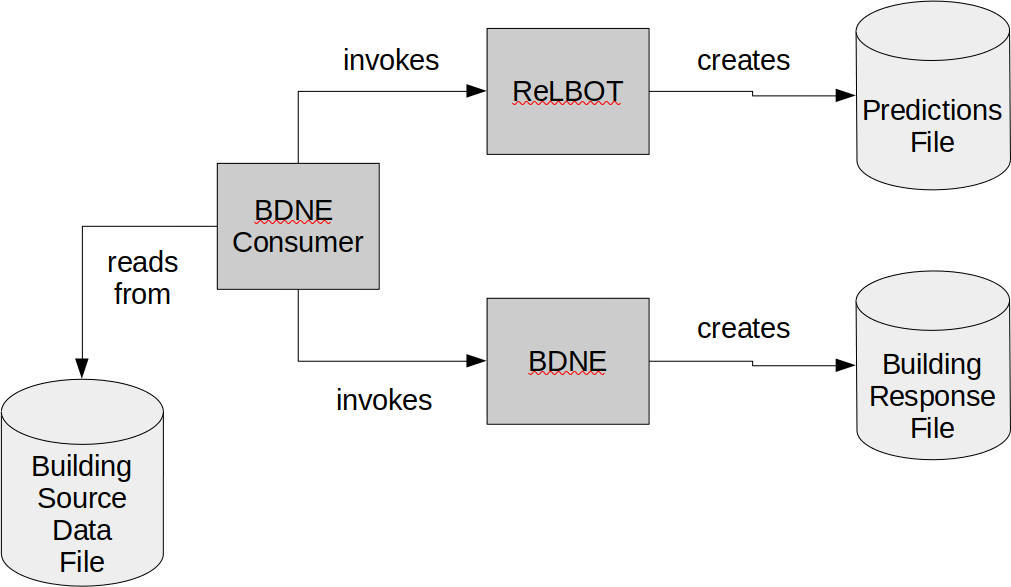}}
\caption{BDNE architecture.}
\label{simulator-architecture}
\end{figure}

The BDNE architecture is shown in Fig. ~\ref{simulator-architecture}. It consists of two components:

\begin{enumerate}
\item The \textit{Building Source Data File}. This CSV file contains actual building sensor data. These data are used as input to the \textit{Streaming Consumer} component.
\item \textit{The BDNE Consumer}. This component reads data from the \textit{Building Source Data File} one row at a time, and invokes \textit{ReLBOT}. ReLBOT takes the row data, which represents the current building state, as input and predicts the action that should be taken next. ReLBOT generates a \textit{Predictions File} that records the action taken for each step, and the reward predicted for that action by ReLBOT.
\item The \textit{BDNE} component emulates the action predicted by \textit{ReLBOT} and adjusts the values read from the \textit(Building Source Data File) by amounts predicted by the BDNE models. The change in values is assumed to be instantaneous. The BDNE generates the \textit{Building Response File}. This file stores that building state values adjusted by BDNE for every step (row) read in from the \textit{Building Source Data File}.
\end{enumerate}

The BDNE implements five machine learning models, one to predict each of the factors described in the sub-section ~\ref{relbot-theory} to calculate the COP. Each model is used to perform regression operations on source data used for Coefficient Of Performance (COP) calculations. These columns are adjusted with new values predicted by the models in response to actions taken by ReLBOT. Columns containing data that would not be impacted by these actions, such as the column containing the outdoor air temperature, are not changed.

Machine learning models used by the BDNE are Artificial Neural Networks (ANNs). Each ANN has logic to adjust the size of the input layer to the size of the feature vector, and 2 identically-sized hidden layers with sigmoid activations. Each model has an output layer with linear activation function.

\subsection{ReLBOT Architecture} \label{relbot-architecture}

ReLBOT architecture is shown in Figure ~\ref{relbot-architecture}. ReLBOT is an actor-critic deep reinforcement learning agent capable of operating with and without reinforcement learning capabilities. When the reinforcement learning feature is turned off, it operates like a standard deep reinforcement learning actor-critic agent.

\begin{figure}[htbp]
\centerline{\includegraphics[width=3.5in]{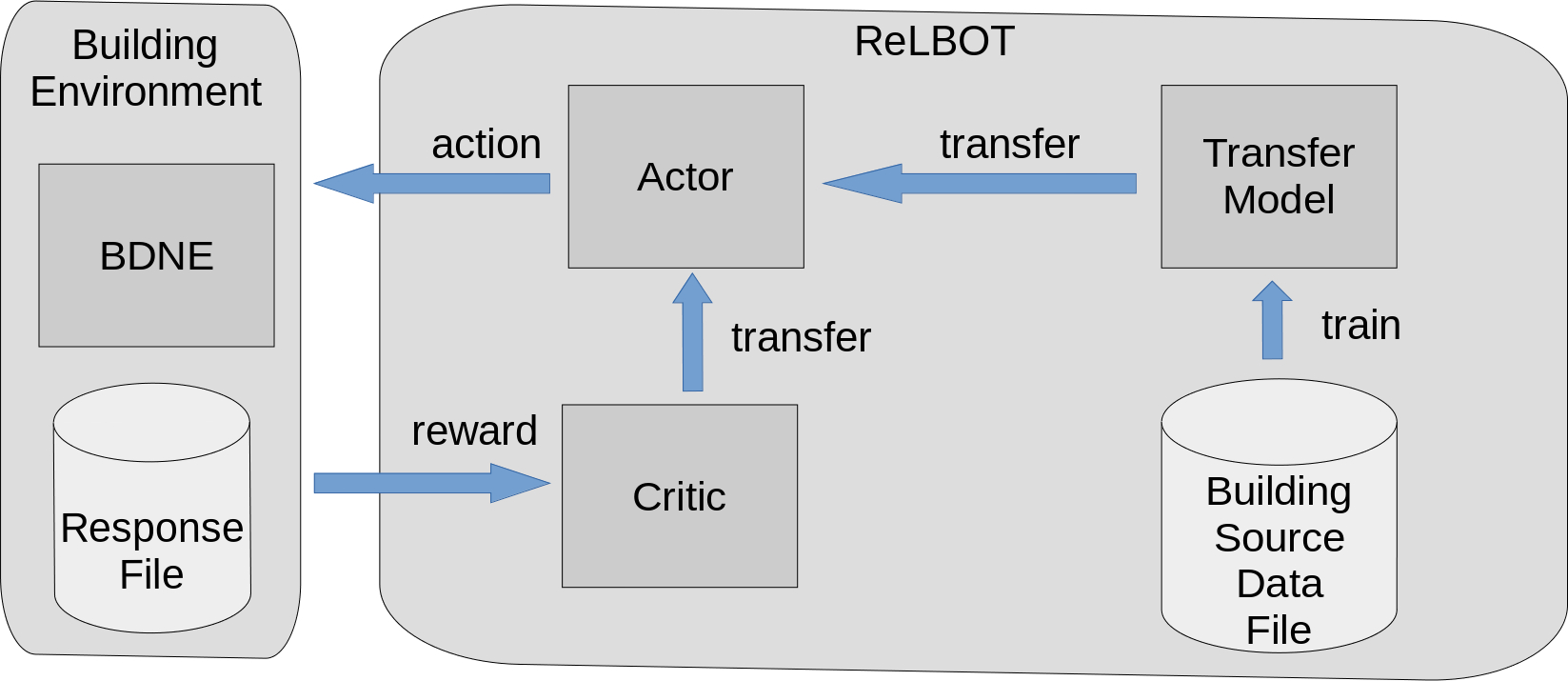}}
\caption{ReLBOT architecture.}
\label{relbot-architecture}
\end{figure}

Both the Actor and the Critic (see Fig. ~\ref{relbot-architecture}) are implemented as ANNs. The role of the Actor is to select the best action given an input target building state. The input target building state is a feature vector containing real numbered values read from the many sensors that instrument the building. The state includes all available values for the building sensors, excluding the timestamp and the chiller set point.

The chiller set point is the value that is being acted on by the ReLBOT actor. At any step the ReLBOT Actor can choose from the following actions ${\{A\}^3_{a=1}}$:

\begin{itemize}
\item Do nothing.
\item Increase the chiller set-point by an amount specified in the ReLBOT configuration file.
\item Decrease the chiller set-point by an amount specified in the ReLBOT configuration file.
\end{itemize}

Once the Actor selects the action based on the current building state, it passes this action to the Building Environment component. The Building Environment component adjusts the target building chiller set point based on the action selected by the ReLBOT actor. The Building Environment component then uses the BDNE to predict the changes in the building state that will result from this action. It will then calculate the actual reward amount that will be returned to the ReLBOT critic component.

The Building Environment component implements a reward function that uses the Coefficient Of Performance (COP) to determine the reward that will be used. The COP is the key building energy performance measure used in the civil engineering field. The amount of reward depends on the difference between the COP calculated for the current target building state, adjusted for changes predicted by the BDNE, and the COP calculated for the previous building state. This difference is then multiplied by a configurable scaling factor.

The Critic (see Fig. ~\ref{relbot-architecture}) predicts the amount of reward ($r_t$) that will be returned by the Building Environment component. Once ReLBOT receives the actual reward returned by the Building Environment component ($R_t$), it uses it as a label ($Y_t$) to re-train the Critic component. The Critic component is incrementally re-trained at every step. Immediately after re-training ReLBOT uses transfer learning to update the Actor ANN using a sub-set of Critic's weights and biases ($\mathcal{T}_s \Rightarrow {\mathcal{T}_t}^c$).

The Actor and Critic ANNs have identical input and hidden layers. The only difference is in the output layer, which in the Actor's case performs a classification task (logistic activation) to identify the best action, and in the Critic's case performs a regression task (linear activation) to predict the reward associated with the action.

\begin{figure}[htbp]
\centerline{\includegraphics[width=3.5in]{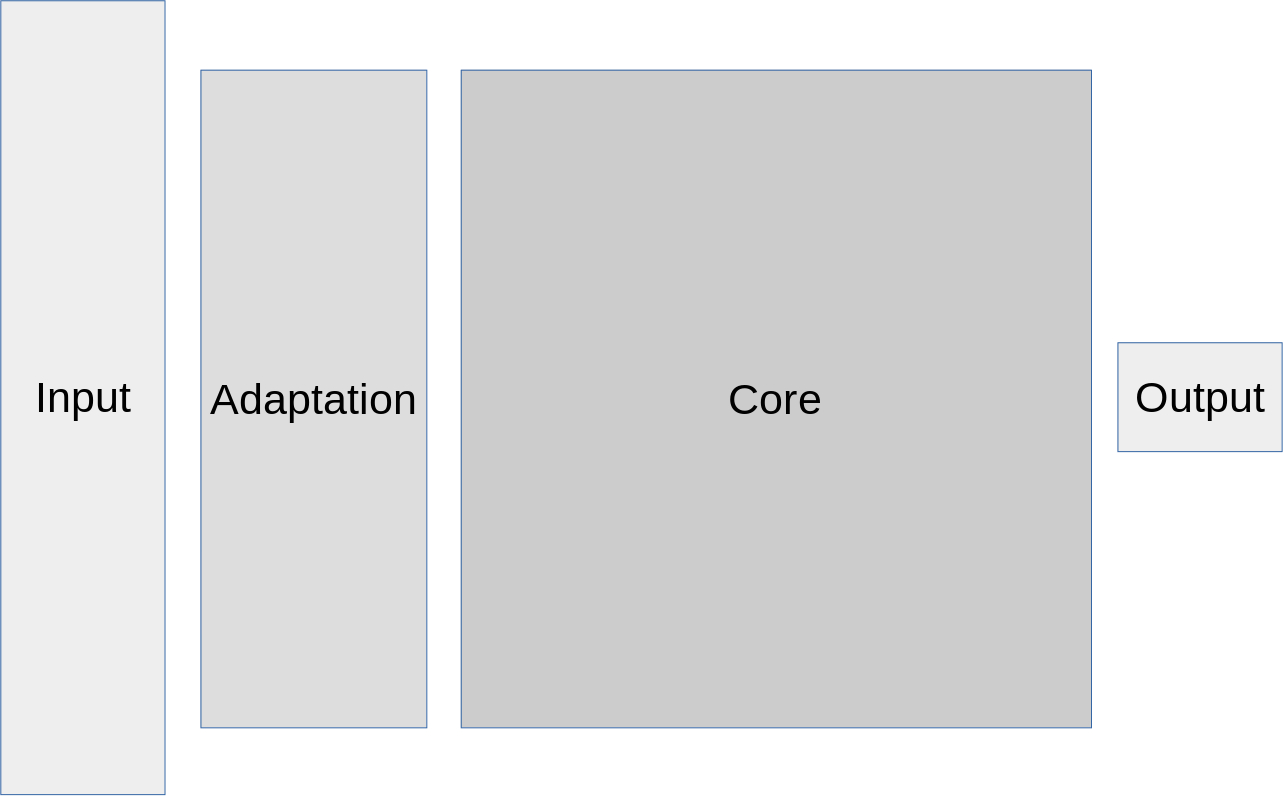}}
\caption{ReLBOT ANN architecture.}
\label{annarch}
\end{figure}

The ANN models used for both the Actor and Critic components have nearly identical architecture. They are both organized into 4 conceptual segments (Figure ~\ref{annarch}):

\begin{itemize}
 \item Input segment. This segment includes a configurable input layer that adjusts to the size of the target building's feature vector and a hidden layer configured to have the same number of neurons as the input layer.
 \item Adaptation segment. This segment includes a hidden layer, with a configurable fixed number of neurons. The number of neurons is the same as in the core segment. The purpose of this layer is to map the variable-size input layer to the fixed size layers of the core segment. This implements the adaptation function $\alpha_t(X_s,X_t)$.
 \item Core segment. This segment has two hidden layers with a configurable, fixed number of neurons.
 \item Output segment. This segment has a single output layer with a single output.
\end{itemize} 

To enable transfer learning ReLBOT uses an off-line utility to read in the transfer building source data file, calculate the COP and the reward for each step in the time-series, and train the \textit{transfer model} for the transfer building. The transfer model is an ANN with an architecture identical to the ReLBOT Critic component model. When the transfer learning feature is turned on ReLBOT uses transfer learning to copy the weights and biases for the Core model segment from the Transfer model to the Actor and Critic models.

\section{Methodology} \label{methodology}

Three buildings located in ASHRAE Climate Zone 5 were selected for our experiments. Table ~\ref{buildings} summarizes the buildings and describes the specifics of their Heating Ventilation and Air Conditioning (HVAC) systems.

\begin{table}[htbp]
\caption{Buildings used as sources of data.}
\begin{center}
\begin{tabularx}{\columnwidth}{XXXXX}
\hline
\textbf{Building ID} & \textbf{Description} & \textbf{Cooling System} & \textbf{Building Age (years)} \textbf{Description} \\
\hline
H & High Efficiency Semi-Hermetic Single-Stage Centrifugal Liquid Chiller with Unit-Mounted VFD  & Variable flow chilled water primary system serving two-pipe fan coils in each unit; variable flow condenser loop served by variable-speed cooling tower & 11 & 24 story, multi-unit residential building with 351 units\\
\hline
T & High Efficiency Semi-Hermetic Single-Stage Centrifugal Liquid Chiller with Unit-Mounted VFD  & Variable flow chilled water primary system serving two-pipe fan coils in each unit; variable flow condenser loop served by variable-speed cooling tower & 30 & 21 story, multi-unit residential building with 200 units\\
\hline
W & High Efficiency Semi-Hermetic Single-Stage Centrifugal Liquid Chiller with Unit-Mounted VFD  & Variable flow chilled water primary system serving two-pipe fan coils in each unit; variable flow condenser loop served by variable-speed cooling tower & 12 & Part of two-tower multi-unit residential complex, each tower with 25-stories, totaling 350 units\\
\hline
\end{tabularx}
\label{buildings}
\end{center}
\end{table}

The following procedure was followed to collect data and evaluate the ReLBOT algorithms:

\begin{enumerate}
\item Data preparation. This involved the following steps:
\begin{enumerate}
\item Remove any identifying information, such as building name or elements of the address, from the raw data file and column name.
\item Impute data for missing values (see discussion above in the ~\ref{bdss-architecture} section).
\item Write out the data in CSV format that could be read in by the BDNE and ReLBOT.
\end{enumerate}
\item Train ReLBOT transfer models for each building.
\item Train BDNE models for each building.
\item Organize experiments by building pairs (target building and transfer building) and execute simulated on-line optimization runs using ReLBOT for each building pair.
\item Aggregate statistics for all runs and metrics (described below).
\end{enumerate}

The following metrics were used to evaluate the effectiveness of ReLBOT:

\begin{itemize}
\item Warm-up reward variance. This metric measures the variance of the reward predicted by the ReLBOT Critic beginning at the very start of the experiment, and until the end of the warm-up period.
\item Mean reward variance. This metric refers to the mean reward variance observed for the entire experiment.
\item Warm-up period duration. The duration of the warm-up period is defined as the number of steps in the building state time-series from the very beginning to the first step where the rolling average of the reward variance is equal to or smaller than the mean variance for the entire time-series recorded in the ReLBOT Predictions File (see Fig. ~\ref{bdss-architecture}). 
\end{itemize}

These metrics were selected for investigation because they provide key insights into the ReLBOT behavior, and the effectiveness of the transfer learning technique. The variance in the reward predicted by the Critic is directly related to the choice of action by the Actor, because the Critic and the Actor ANN models share most of the weights and biases. High variance in the predicted reward is thus related to sub-optimal, and more chaotic choice of action selected by the Actor. The period (number of steps) during which the predicted reward variance is much larger than the overall mean predicted reward variance can be used to unambiguously define the length of the warm-up period for the reinforcement-learning agent. This is graphically shown in Figure ~\ref{t-w-rewards}.

\section{Results} \label{results}

Table ~\ref{summary} presents a summary of findings for our experiments. This table shows relative improvement achieved for the key metrics as times-factor. For all metrics smaller is considered better, and so the times-factor is defined as the metric with transfer learning divided by the corresponding metric without transfer learning. For example for the building combination T-W, where the target building is T and the transfer building is W, the reduction in the duration of the ReLBOT warm-up period was observed to be 2.34 times shorter with transfer learning than without it. The warm-up reward variance was observed to be 247.48 times smaller with transfer learning than without it. The mean variance was observed to be 17.91 times smaller with transfer learning than without it. In this case transfer learning among buildings clearly produced a dramatic improvement to the speed with which ReLBOT was able to find the COP optimum and minimized the potential discomfort that would be experienced by the building residents due to excessive exploration of the action space.  

\begin{table}[htbp]
\caption{Improvements observed with transfer learning.}
\begin{center}
\begin{tabularx}{\columnwidth}{XXXXX}
\hline
\textbf{Target Building} & \textbf{Transfer Building} & \textbf{Warm-up Duration Reduction (times)} & \textbf{Warm-up Variance Reduction (times)} & \textbf{Mean Variance Reduction (times)} \\
\hline
H & T & 4.59 & 4.41 & 2.08\\
T & H & 1.00 & 3.24 & 3.01\\
W & T & 1.00 & 1.02 & 1.02\\
W & H & 1.00 & 1.55 & 1.38\\
T & W & 6.20 & 131.63 & 31.78\\
H & W & 4.45 & 8.03 & 2.99\\
\hline
\textbf{Average} && \textbf{3.04} & \textbf{24.98} & \textbf{7.04} \\
\hline
\end{tabularx}
\label{summary}
\end{center}
\end{table}

Averaged for all six building combinations, it was observed that transfer learning among buildings resulted in:

\begin{itemize}
\item 3.04 times reduction in the duration of the warm-up period.
\item 24.98 times reduction in reward variance during the warm-up period.
\item 7.04 times reduction in mean reward variance. 
\end{itemize}

It should be note that in some cases transfer learning among buildings does not seem to produce significant improvements in the key metrics. For example for building combination W-T (W as the target building and T as the transfer building, see Table ~\ref{summary}), using transfer learning did not make many difference. 

For the building combination H-T some of the metrics were observed to degrade with transfer learning. The warm-up variance increased significantly and the mean variance incresed slightly.

\begin{figure}[htbp]
\centerline{\includegraphics[width=3.5in]{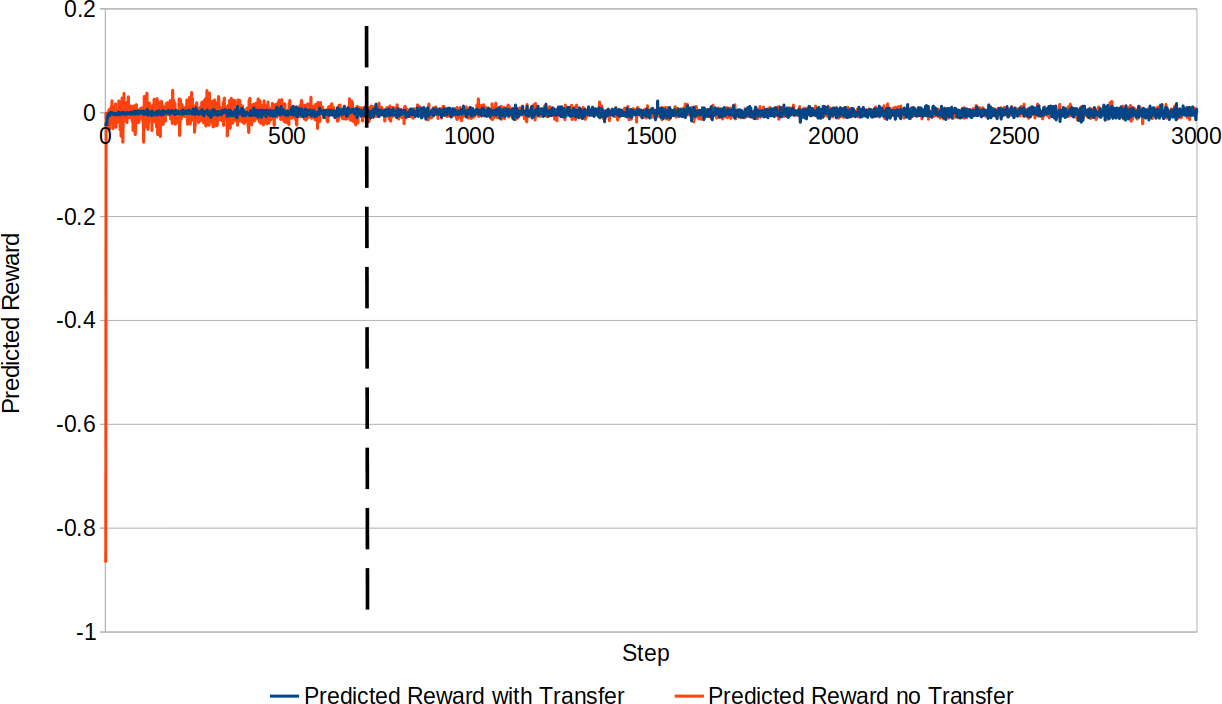}}
\caption{Predicted reward with and without transfer learning for building combination T-W.}
\label{t-w-rewards}
\end{figure}

Figure ~\ref{t-w-rewards} shows the ReLBOT predicted reward behavior with and without transfer learning. Without transfer learning the predicted reward varies widely during the warm-up period, the duration of which is indicated on the graph by the dashed line. During this period of time, because the ReLBOT Actor and Critic share knowledge, the Actor explores the action space almost at random. Since each action results in a change to the chiller set point this could cause considerable discomfort to the building residents. In this case the warm-up period lasts for close to three weeks.

With transfer learning from building W, the warm-up period was observed to be almost completely eliminated, reducing the potential for discomfort to the building residents. In both cases the predicted reward eventually tends to zero, as the ReLBOT finds the COP optimum for the target building.

\begin{figure}[htbp]
\centerline{\includegraphics[width=3.5in]{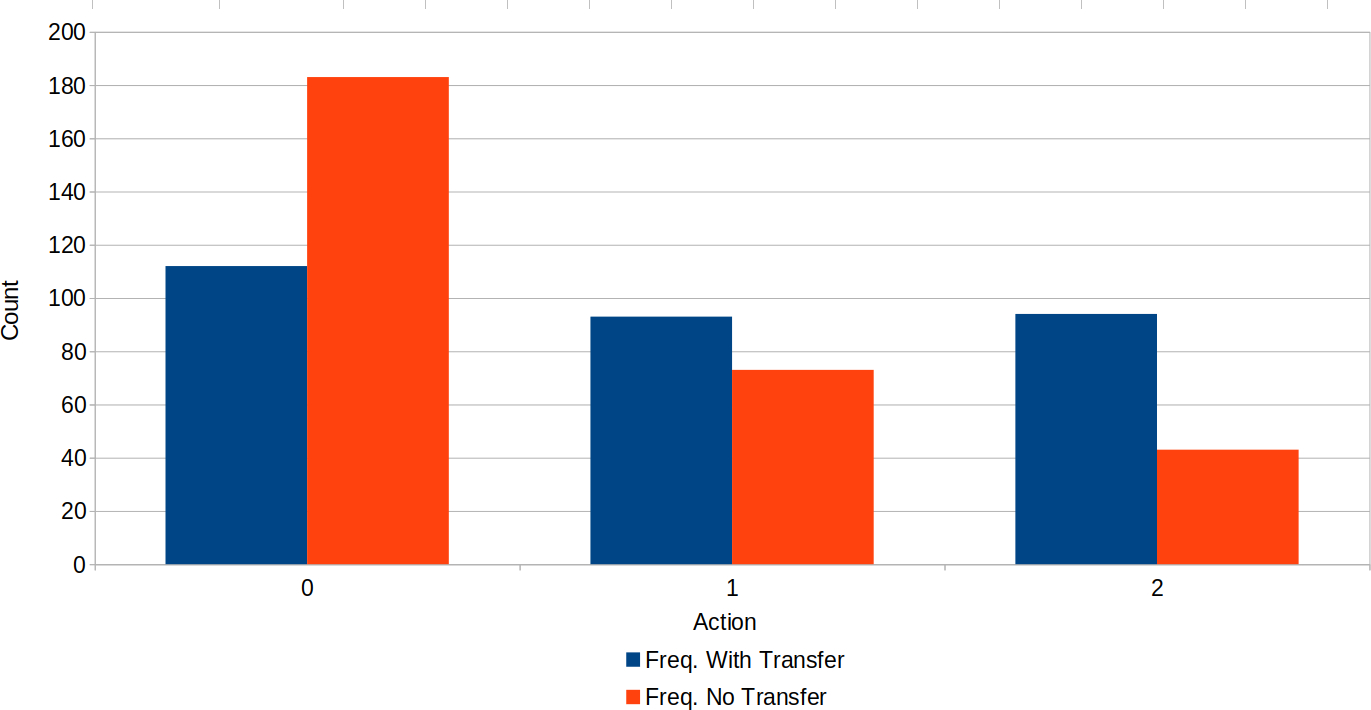}}
\caption{Frequency of actions taken by ReLBOT with and without transfer learning for building combination T-W during the warm-up period (first 300 steps taken).}
\label{actions}
\end{figure}

Figure ~\ref{actions} shows the distribution of actions taken by ReLBOT during the warm-up period. Because the duration of the warm-up period varies with and without transfer learning, the first 300 steps were used. 

With transfer learning it was observed that the actions chosen by the ReLBOT during the warm-up period were much more evenly distributed between choosing to do nothing, and adjusting the chiller set point up or down to find the optimal operating conditions. This pattern of behavior leads to more gradual adjustments to the chiller set-point, reducing the potential to cause discomfort to the building residents.

\begin{figure}[htbp]
\centerline{\includegraphics[width=3.5in]{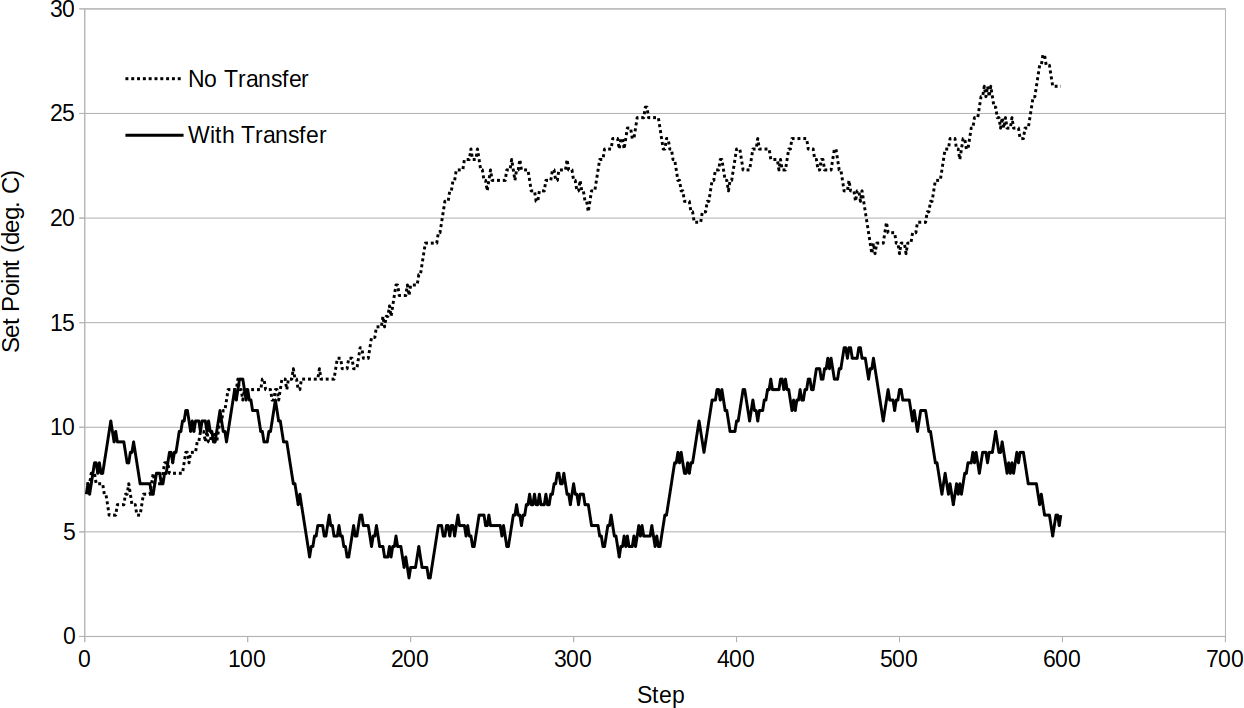}}
\caption{Set point behavior with and without transfer learning during the warm-up period for building combination T-W.}
\label{setpoints}
\end{figure}

Figure ~\ref{setpoints} demonstrates this behavior for the building combination T-W with and without transfer learning during the warm-up period. Without transfer learning the ReLBOT actor-critic RL agent aggressively adjusts the chiller set point to higher and higher values in search of the COP optimum. Effectively, it turns off the cooling by pushing the chiller set point above 20 degrees Celsius. If this pattern of set points was enacted on an actual building, it would cause significant discomfort to the building residents.

In comparison, with transfer learning, the agent adjusts the set point much more conservatively, exploring values slightly above and below the initial set-point of 7.3 degrees Celsius. This patter of chiller set-points would not cause significant discomfort to the building residents. 

ReLBOT was eventually able to find the same COP optimum both with and without transfer learning. Without transfer learning though, it took longer to get there and chiller set-point fluctuations were more significant. The primary benefit of using transfer learning from another building is the fact that it mitigates the risk of causing significant discomfort to the building residents during initial building commissioning, or during on-going re-commissioning as part of SOCx. An added benefit are the reduced energy cost expenditures that result from the much-shortened warm-up period.

It is important to note that the effectiveness of transfer learning among buildings depends on the choice of the transfer building. In our case building W appears to be the best transfer building, producing the most spectacular results with two different target buildings - T and H. 

\begin{figure}[htbp]
\centerline{\includegraphics[width=3.5in]{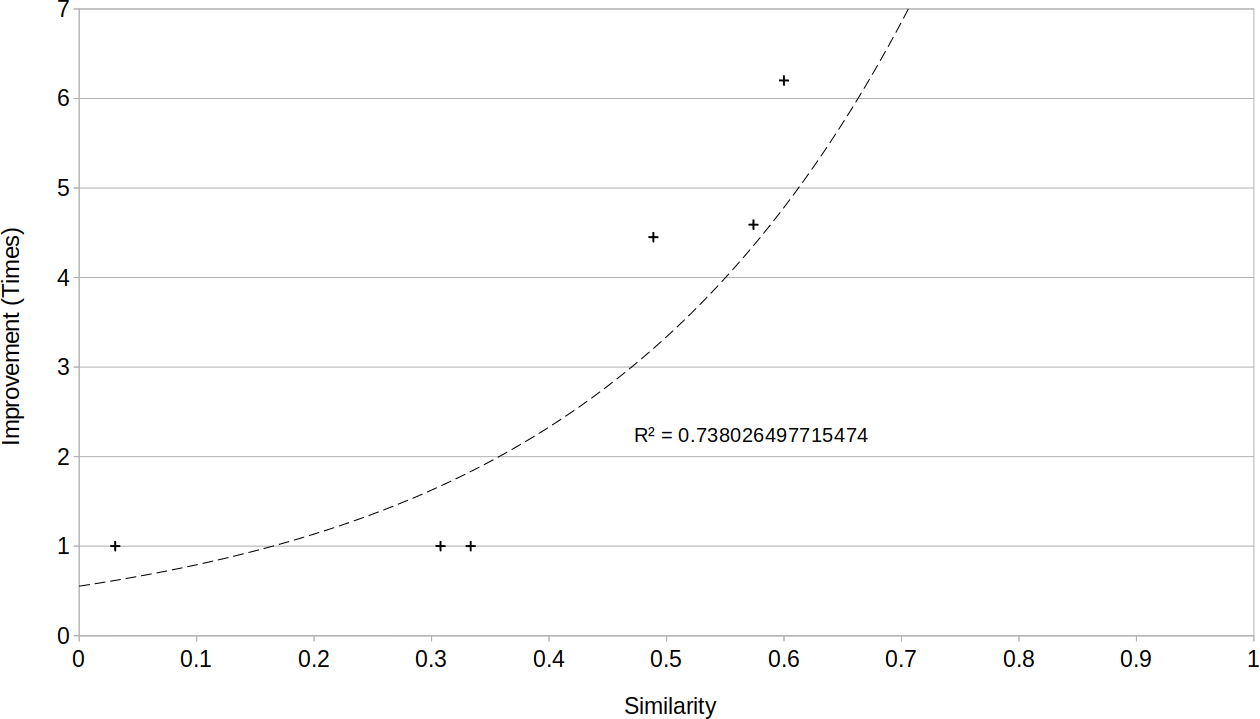}}
\caption{Improvement (times) vs. Similarity plotted for the warm-up duration reduction.}
\label{wulr-similarity}
\end{figure}

\begin{figure}[htbp]
\centerline{\includegraphics[width=3.5in]{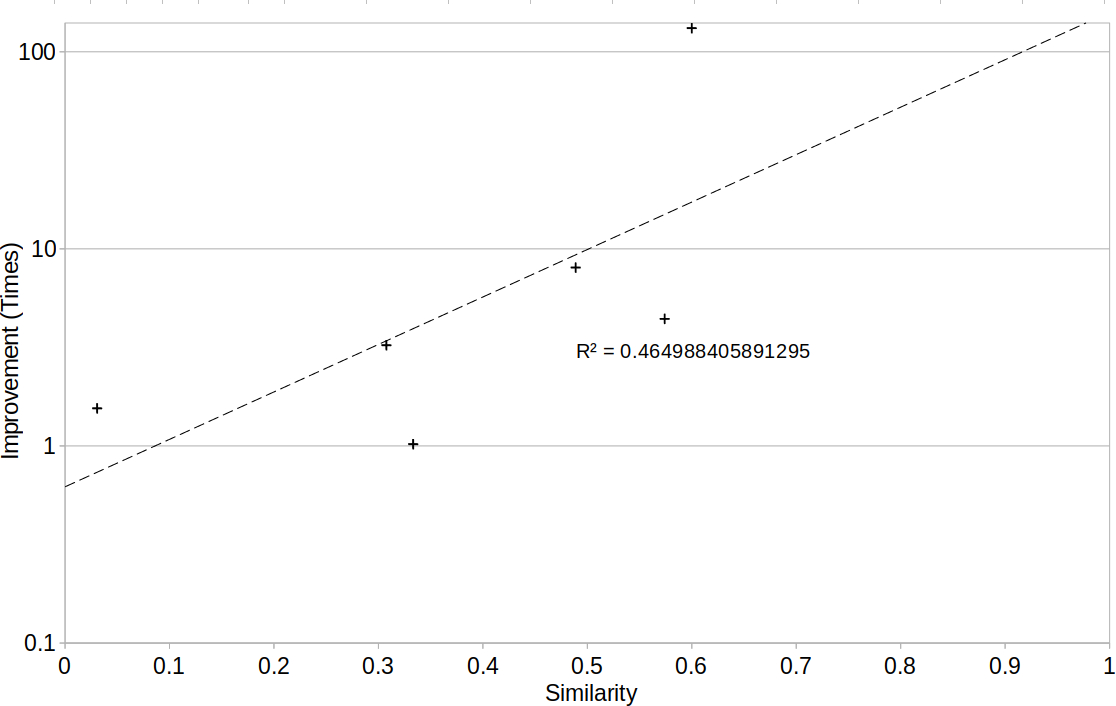}}
\caption{Improvement (times) vs. Similarity plotted for the warm-up variance reduction.}
\label{wuvr-similarity}
\end{figure}

\begin{figure}[htbp]
\centerline{\includegraphics[width=3.5in]{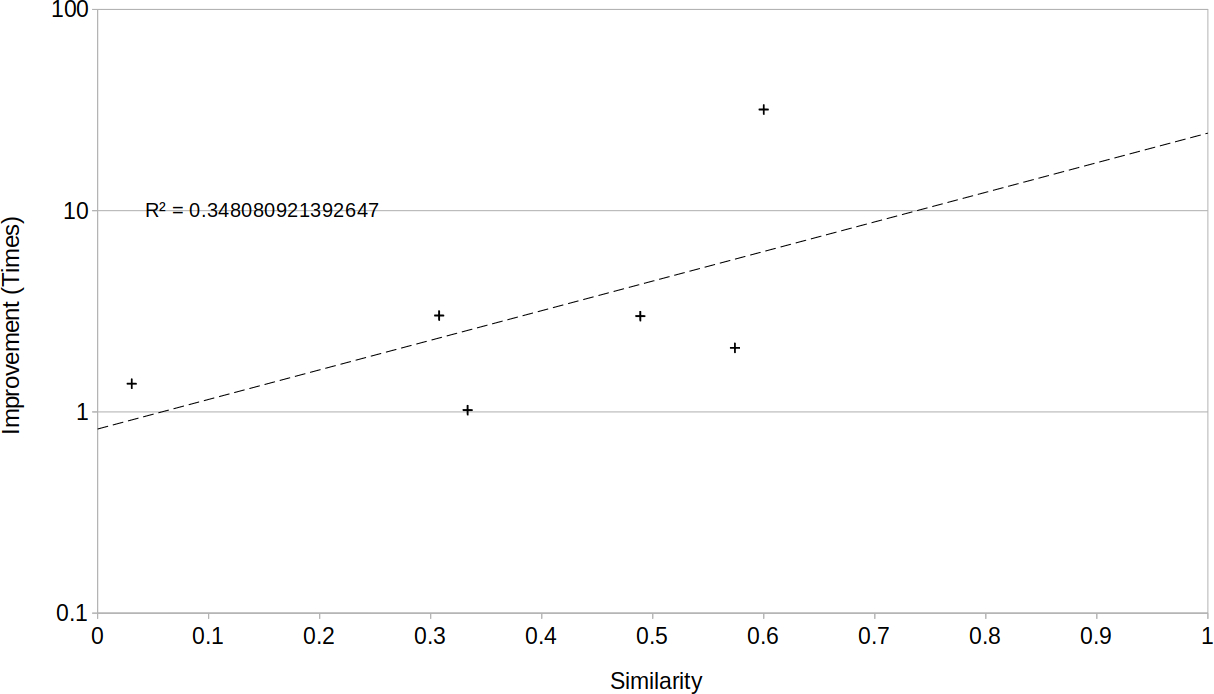}}
\caption{Improvement (times) vs. Similarity plotted for the mean variance reduction.}
\label{mvr-similarity}
\end{figure}

Figures ~\ref{wulr-similarity}, ~\ref{wuvr-similarity}, and ~\ref{mvr-similarity} show the relationship between the amount of improvement (times improvement) and similarity for all of the key metrics. The overall trend in the data is shown using a dashed line. Note that for the Mean Variance Reduction (MVR) and Warm-Up period Variance Reduction (WUVR) the y-axis is logarithmic. 

It should be noted that the feature vectors of our buildings were observed to be relatively dissimilar (i.e. relatively far from the ideal similarity of 1). Similarity for our building combinations ranges from 0.03 (W-H) to 0.60 (T-W). For all metrics the overall trend shows significant improvement over this range, and the relationship appears to be exponential, with order-of-magnitude improvements achieved as similarity reaches 0.6.

For warm-up duration reduction, the coefficient of determination $R^2$ for the exponential model was observed to be 0.74 (see Figure ~\ref{wulr-similarity}), supporting the notion that the relationship between similarity and this metric is exponential. For WUVR and MVR $R^2$ for the exponential model was observed to be lower at 0.46 (see Figure ~\ref{wuvr-similarity}) and 0.35 (see Figure ~\ref{mvr-similarity}) respectively. This was observed to be due to a single data point (building combination T-W) that resulted in better-than-exponential improvement for WUVR and MVR. Overall, the exponential model best explains the observed relationship between similarity and our key performance metrics.

The best results were observed in those cases where the length of the transfer building feature vector was smaller than the length of the target building feature vector, and most transfer building features were statistically similar to target building features.

The worst results were observed in those cases where the length of the transfer building feature vector was larger than the length of the target building feature vector. This appears to result in only partial knowledge transfer, and reduced benefit from transfer learning.

\section{Conclusion} \label{conclusions}

In this work we presented ReLBOT - the first reinforcement learning technique that allows smart buildings to overcome the ’cold start’ scenario by transferring knowledge from another, already optimized building. The building used as the transfer building does not have to be a smart building, but it does need to have a sufficient level of instrumentation and a historical energy performance data set to enable training of transfer models.

Our technique uses deep reinforcement learning in combination with transfer learning to greatly mitigate the optimization algorithm deployment risk to the building management company. ReLBOT can reduce the duration of the warm-up period by more than 6 times (3 times on average), reduce the variance observed during the warm-up period by up to 132 times (25 times on average), and reduce the overall mean variance by up to 32 times (7 times on average).

For optimal results the transfer building needs to be carefully selected to ensure that its feature vector is the same size, or smaller than the feature vector of the target building. The feature vector of the target building should ideally contain statistically equivalent features to all of the features of the transfer building, if sufficient data are available to make this comparison. For this reason, the use of buildings with rich data sets offer significant benefits for transfer learning. However, even relatively dissimilar building feature vectors, can still lead to dramatic improvements in all key warm-up period metrics, as well as a significant reduction of risk associated with RL deployment and, thus, have significant value for energy management.

Limitations of this study are the relatively small (4) number of buildings considered, and the consideration only of cooling system (chiller) performance. To address these limitations, future work includes extending ReLBOT to optimize building performance during the heating cycle (boiler efficiency), as well as adding more buildings of varying topologies, HVAC systems, and levels of sensor instrumentation. Other future work of value would be to further expand this across climate zones, integrating the methods developed by \cite{ribeiro2018transfer}, and to integrate RelBot with smart commissioning approaches to support semi-autonomous building start-up and energy optimization.

\section{Acknowledgment} \label{acknowledgement}

This research was funded by the Natural Science and Engineering Research Council of Canada [DGECR-2018-00395 and RGPIN/04105-2018].



\bibliographystyle{elsarticle-harv} 
\bibliography{genkin-mcarthur-relbot}

\end{document}